# FSD: Feature Skyscraper Detector for Stem End and Blossom End of Navel Orange †

SUN Xiaoye[1,2], LI Gongyan[1] and XU Shaoyun[1]

1. Institute of Microelectronics of Chinese Academy of Sciences, Beijing 100029, China
2. University of Chinese Academy of Sciences, Beijing 100049, China

sunxiaoye16@mails.ucas.ac.cn, {xushaoyun, ligongyan}@ime.ac.cn

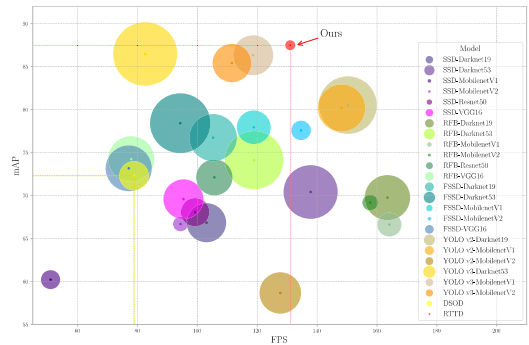

Fig. 1 **FSD** delivers the best performance with the most compact architecture and faster speed in our navel orange dataset. The size of the circle represents the size of the model.

**Abstract** — To accurately and efficiently distinguish the stem end and the blossom end of navel orange from its black spots, we propose a feature skyscraper detector (FSD) with low computational cost, compact architecture and high detection accuracy. The main part of the detector is inspired from small object that stem (blossom) end is complex and black spot is densely distributed, so we design the feature skyscraper networks (FSN) based on dense connectivity. In particular, FSN is distinguished from regular feature pyramids, and which provides more intensive detection of high-level features. Then we design the backbone of the FSD based on attention mechanism and dense block for better feature extraction to the FSN. In addition, the architecture of the detector is also added Swish to further improve the accuracy. And we create a dataset in Pascal VOC format annotated three types of detection targets the stem end, the blossom end and the black spot. Experimental results on our orange data set confirm that FSD has competitive results to the state-of-the-art one-stage detectors like SSD, DSOD, YOLOv2, YOLOv3, RFB and FSSD, and it achieves 87.479%mAP at 131 FPS with only 5.812M parameters.

**Keywords** — Real-time Small Object Detection Convolutional Neural Network, Feature Skyscraper, Navel Orange

## I. Introduction

The surface defect of navel orange can be easily detected in traditional image processing, but the stem end and the blossom end of the navel orange are also drastically mistaken as defect. This may drive up economic losses. Despite the four types of symptoms hard spot, false melanoses or speckled blotch, freckle spot, and virulent or spreading spot commonly appear as black spots or blotches[1], we still only focus on the visual black spot, which is easier to confuse with a stem end and a blossom end. With the breakthrough progress in deep learning in image processing in recent years, that it surpasses the performance of traditional image processing[2] allows us to apply this technique to reduce the false positive rate.

Most of the models for detection tasks today are based on public data sets, like ImageNet, MS COCO, Pascal VOC, CIFAR-10, etc. Their good versatility often requires a more complex architecture. For some specific data sets with few detection categories and high homogeneity, although the models may have good performance after fine tuning, such architectures are very redundant and takes up too much computational cost. Especially for the navel orange detection task, the data set is relatively simple and has a strong distribution law. Therefore, we consider whether we can design and optimize the network architecture based on the statistical characteristics of the dataset, so that the network architecture can achieve high performance with the most compact structure possible. Indeed, our model FSD shows good performance as shown in Fig. 1.

Fine-tuning restricts the design of the model architecture, and may cause learning bias and domain mismatch problems[3], so we adopt the dense connectivity proposed in DenseNet[4]. This strategy makes training from scratch in DSOD[3] possible.

Different from the regular feature pyramids, we design the feature skyscraper in our detector, as shown in Fig. 2. It is primarily inspired by the statistical result that the navel orange dataset detection object is small and complex. And this design decouples the dependence of the multi-scale features on the resolutions of the feature maps, and achieves the multi-scale features by presetting the default boxes. The resolution of each layer of the skyscraper remains the same so that the higher level can design smaller default boxes. However, the higher levels of the general feature pyramids are invalid for small objects. This designed structure ensures the invariance of the resolution of the feature map, and avoids the problem that the low-resolution feature map limits the design of the minimum default box. Furthermore, the default box corresponding to the feature map can effectively cover the detection area at each scale, and larger detection targets also have denser default boxes. The size and aspect ratio of the default boxes corresponding to the feature map are determined by the bounding boxes of clustering our navel orange dataset.

† This work is supported by National Key R&D Program of China (No. 2018YFD0700300).

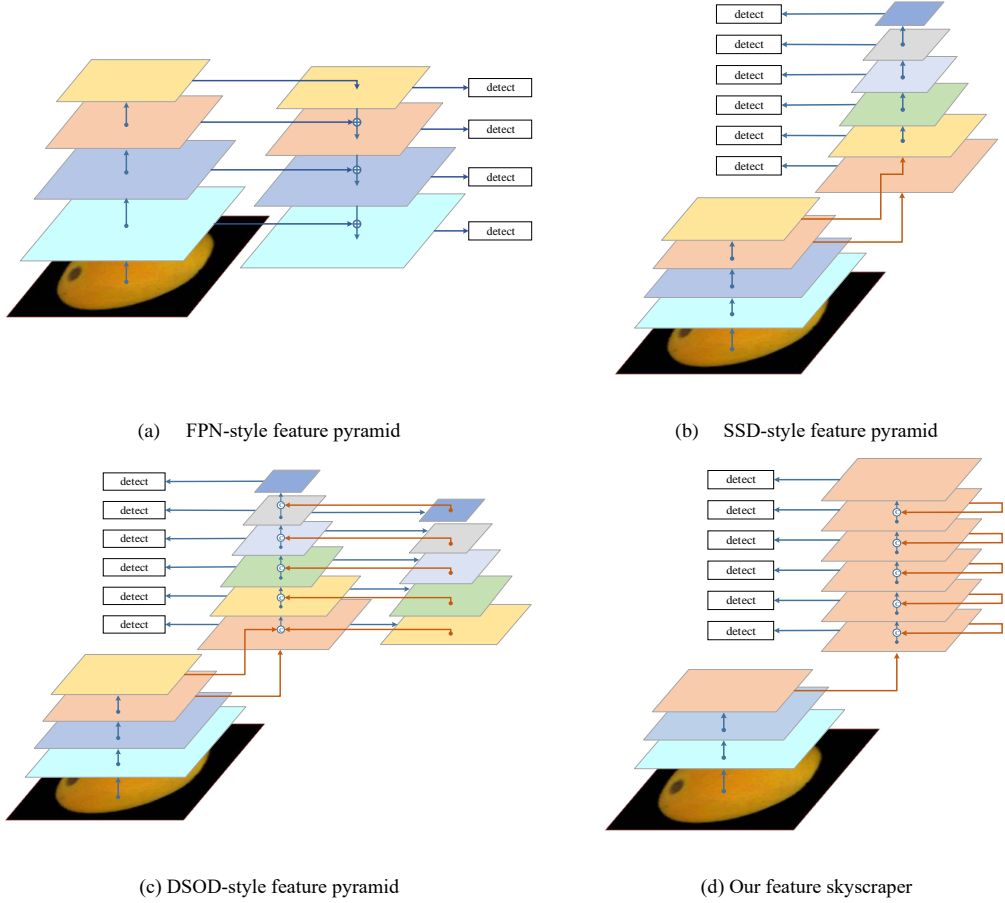

(a) FPN-style feature pyramid  (b) SSD-style feature pyramid

(c) DSOD-style feature pyramid  (d) Our feature skyscraper

Fig. 2 Feature skyscraper and several feature pyramids

For the backbone, it consists of three parts: stem, dense block, and transition layer. Dense block is similar to FSN and is designed by dense connectivity. The transition layer connects the dense block and the feature skyscraper network. We add an Squeeze-and-Excitation (SE) layer[5] to both the stem and transition modules. And all activation functions in the architecture use the Swish[6], except that SE layer use the sigmoid function.

As shown in Fig. 1, we compare performance of our model with some state-of-the-art and classical one-stage models, such as SSD[7], DSOD[3], YOLOv2[8], YOLOv3[9], RFB[10] and FSSD[11]. And we also experiment with the performance of these models under different backbones, such as DarkNet19[8], DarkNet53[9], MobileNetV1[12], MobileNetV2[13], ResNet50[14], VGG16[15]. The experimental results confirm that the FSD is the most compact architecture for the best performance in real time. And the feature skyscraper network designed and trained based on dataset statistics have significant advantages. Finally, we evaluate the effectiveness of adding the SE layer and using the Swish activation function, and experimentally explore the most compact architecture. The main contributions in this paper:

(1) We build a navel orange dataset to detect the stem end and the blossom end from the black spot.
(2) We propose feature skyscraper detector (FSD), which is optimized based on the statistical results of the data set.
(3) We design the feature skyscraper network which is better for feature pyramid in our data set.
(4) We show that FSD can implement state-of-the-art performance on our navel orange dataset and verify the effectiveness of the optimization for its architecture.

This paper first introduces the detection of surface defect in navel oranges and some of the current detection techniques in Section II. Then in Section III, the basics of the image acquisition device, the navel orange data set, and the statistical results for it are shown. Then, Section IV shows the architecture we proposed and the settings for its training. In Section V, we present the details of the model implementation, compare it with the current state-of-the-art models. Next, in Section VI verifies the validity of the design and improvement of the model. The final section VII concludes this paper.

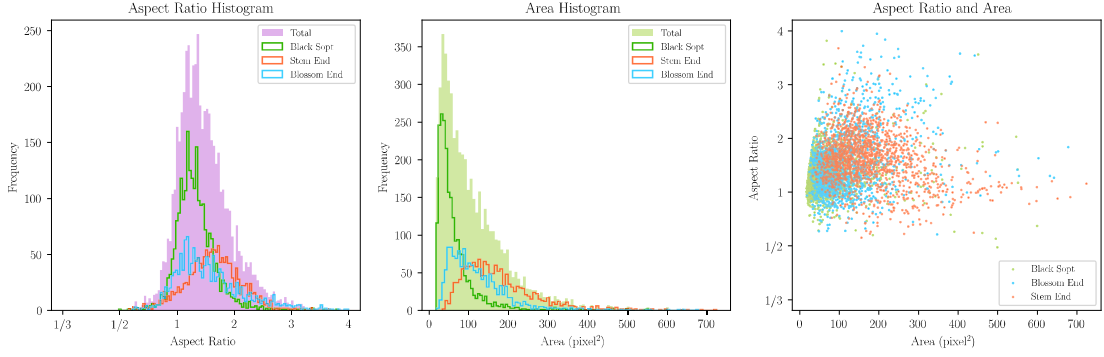

Fig. 3 Histograms of the distribution of the area size and aspect ratio of the bounding boxes

## II. Related Work

Before the deep learning explosion, the orange peel defects detection task is generally implemented by machine learning. Behera et al.[16] classify the orange disease and compute its severity by SVM with K-means clustering and Fuzzy logic. Rong et al.[17] proposes an adaptive lightness correction algorithm to solve the problem that the uneven distribution of lightness on the surface of the navel orange is difficult to detect in the dark region. However, the performance and generalization of these methods are not good, and the design of the model depends on the environment. Zhang et al.[18] identify the apple's bruises and blemishes from the stem end and calyx of apple images by near-infrared spectrum light, but it's hardware cost is relatively high.

The study by Kamilaris et al.[2] shows that deep learning algorithms are indeed superior in accuracy to existing used image processing techniques. So, we consider the current detection architecture and some techniques in deep learning to design our model based on statistics of dataset.

**Detection Methods** Since the deep learning boom, there are mainly two types of detectors, the Two stage detection framework and the One stage detection framework[19, 20]. Among them, because it is a multistage complex pipeline, the training process of the two stage detection framework (Representative RCNN[21], Fast RCNN[22], Faster RCNN[23] and RFCN[24] etc.) is more complicated, the optimization is difficult, and the time of inference is very slow[19]. In comparison, one stage detection represented by YOLO[25], and SSD[7], etc., although achieving relatively low object detection quality, can avoid the problems mentioned above.

YOLO[25] puts the detection problem as a regression problem and directly outputs the relevant information of the entire image. SSD[7] applies multi-scale feature maps for detection at multi-scales, and uses small convolutional filters applied to feature maps to get information about category and location.

**Backbones** It is critical for backbones in the object detection task[26]. Representative backbones include AlexNet[27], VGG[15], Inception series[28-30], ResNet[14], SENet[5], DenseNet[4], MobileNet series[12, 13], and so on. Although many backbones are designed for classification, as the classification performance increases, the performance of the object detection is also improved[26, 31]. Due to the superiority of dense connectivity we mainly consider the DenseNet.

**Fusion of layers** It is typical for **feature pyramid** to integrate information from different feature map[20]. Yi Lin et al. [32]propose the feature pyramid network to fuse the feature maps. And FSSD[11] designs a structure to make it easier to fuse the feature maps from different scales. M2Det[33] introduces more complex multi-level multi-scale features to detect complex small object. But for simple detection tasks, these pyramid feature fuses make the model more complicated and training difficult. Instead, we apply **dense connectivity**, and it is introduced in DSOS[3], which makes it inherit the features of DenseNet[4]'s training from scratch. The connection method is derived from ResNet[14], but it helps to better propagate features and losses and reduce the number of model parameters[4].

**Channel-wise Attention Mechanism** Squeeze and Excitation[5] is one of channel attention mechanism, which can improve performance by a flexible way and only need a few additional computational cost[19]. Applying it, both CliqueNet[34] and M2Det[33] get a performance boost.

**Activation Functions** Ramachandran et al.[6] discovered novel activation functions named Swish by automatic search techniques. And compared to the usual activation functions such as ReLU[35], PReLU[36], ELU[37] and SELU[38] etc., they confirm that Swish has the best empirical performance. This was further confirmed in our model FSD.

## III. Background

In this section, first describe how we acquire the images by the acquisition device, as well as some of its parameters and task requirements. Then introduce the construction of the dataset and the statistical characteristics of the dataset.

## 1. Image acquisition

The experimental RGB color images are collected at resolution 1280×1024 from a navel orange grading machine[2], as shown in Fig. 4.The machine vision part is mainly composed of high-resolution industrial cameras above the conveyor belt with rollers, LED warm light sources for providing sufficient light to the camera, and photoelectric switch for controlling image capture.

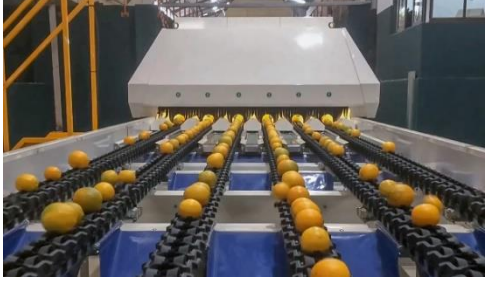

Fig. 4 The high-performance orange grading machine.

The navel orange triggers the photoelectric switch to capture images by the camera which rate up to 16 frames per second. At the same time, with the roller to drive the navel orange rotation, the camera can capture a series of images at different angles of an orange. After passing through the machine vision device, navel oranges on the orbit are popped out by the spring device to classify according to the detection results of the machine vision part.

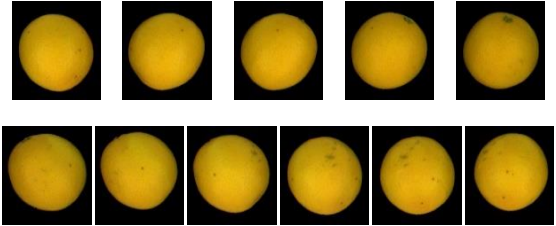

Fig. 5 Navel orange image collected and preprocessed by the device

The machine vision part can obtain 11 different angles images of each navel orange to ensure that sufficient surface information is provided. And after preprocessing these images, the orange region of each image is extracted. As shown in Fig. 5, they are the same height and different width 11 images. And in order to meet the requirement of system real-time detection, the time to get result from the detection algorithm is less than 9ms per image(111fps).

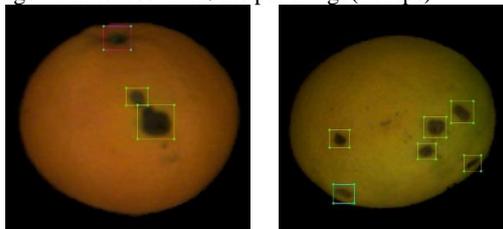

(a) The red one annotated the stem end  (b) The blue one annotated the blossom end

Fig. 6 The detection object is annotated with a bounding box.

Those yellow bounding boxes are black spots.

## 2. Data Set

We collected 11187 images from the machine vision part. Each image is required to be marked with the stem end, the blossom end and the black spot in Pascal VOC format. In order to balance every detection object of data set, we extract the dataset based on the minimal mark of category.

There are 3,482 color images of navel oranges labeled with 1,583 stem ends, 1,482 blossom ends and 2,250 black spots as experimental data sets. Each detection object is marked with a bounding box as shown in Fig. 6. One tenth of these images (348) are randomly selected as the test set, and the remaining images (3134) are used as training sets. Table 1 gives more details about this data set.

Table 1 The navel orange dataset

|  | Train | Test | Total |
| --- | --- | --- | --- |
| Stem end | 1,418 | 165 | 1,583 |
| Blossom end | 1,335 | 147 | 1,482 |
| Black spot | 2,012 | 238 | 2,250 |
| Total | 4,765 | 550 | 5,315 |
| images | 3,134 | 348 | 3,482 |

## 3. Statistics

Since the input to our architecture is $150\times150$, the images with bounding boxes are resized to $150\times150$ resolution before statistics. Then we get the width and height information of all bounding boxes in the data set. As shown in Table 2, the width-height aspect ratio and area size are then calculated, because the architecture of our model and the selection of the default boxes are related to these two statistics in the subsequent experiments. Next, further visualize these information and results, as shown in Fig. 3.

Table 2 Bounding boxes statistics

|  | W(pix) | H(pix) | Aspect | Area(pix$^2$) |
| --- | --- | --- | --- | --- |
| Mean | 12.130 | 8.389 | 1.483 | 113.095 |
| Std | 5.307 | 3.420 | 0.478 | 88.785 |
| Min | 4.054 | 4.008 | 0.494 | 16.858 |
| Max | 40.057 | 31.797 | 3.996 | 724.138 |

**Area size**  It is obvious that the area of bounding boxes is generally small in Fig. 3. Compared to SSD and DSOD, the scale setting is 0.2 to 0.9, whereas the scale range of bounding boxes is (0.02672, 0.267) (Calculated by Min(W,H)/150 and Max(W,H)/150). And according to the Table 2 results, the maximum area size is only 3.218% of the full image. This implies that the architectural design of our model should focus more on small detection objects and the setting range of the default box size can be further reduced.

**Aspect ratios**  It is worth noting that the aspect ratio is not symmetrical distribution in 1 nearby and its distribution is also relatively concentrated as shown in Fig. 3. This implies that the selection of the aspect

---

[2] This machine is jointly developed by Jiangxi Reemoon Sorting Equipment Co., Ltd. and Institute of Microelectronics of Chinese Academy of Sciences.

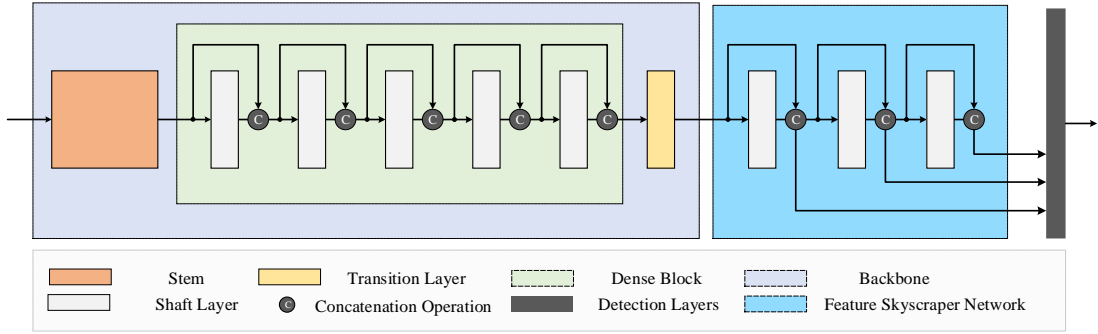

Fig. 7 Architecture of FSD.

Table 3 FSD architecture configuration. The growth rate for dense block is $k=48$, and for FSN is $k=144,96,28$. Each shaft layer corresponds the sequence BN-swish-Conv and each convolution layer is the sequence Conv-BN-swish.

| Module | | Output Size (Input 3×150×150) | FSD |
|---|---|---|---|
| Stem | Convolution Layer | 48×150×150 | $3\times3, 48$ conv, stride 1, padding 1 |
| | Convolution Layer | 96×150×150 | $3\times3, 96$ conv, stride 1, padding 1 |
| | SE Layer | 96×150×150 | global avg pool $[96,32,96]\ fc$ |
| | Pooling Layer | 96×75×75 | $2\times2$ max pool, stride 2 |
| Dense Block | Shaft Layer 1 | 144×75×75 | $\begin{bmatrix} 1\times1\ conv \\ 3\times3,\ conv,\ padding\ 1 \end{bmatrix} \times 5$ |
| | Shaft Layer 2 | 192×75×75 | |
| | Shaft Layer 3 | 240×75×75 | |
| | Shaft Layer 4 | 288×75×75 | |
| | Shaft Layer 5 | 336×75×75 | |
| Transition Layers | Convolution Layer | 336×75×75 | $1\times1, 336$ conv, stride 1 |
| | SE Layer | 336×75×75 | global avg pool $[336,168,336]\ fc$ |
| | Pooling Layer | 336×38×38 | $2\times2$ max pool, stride 2 |
| Feature Skyscraper Network (FSN) | Shaft Layer 1 | 480×38×38 | $1\times1, 480$ conv $3\times3, 480$ conv, stride 1, padding 1 |
| | Shaft Layer 2 | 576×38×38 | $1\times1, 576$ conv $3\times3, 576$ conv, stride 1, padding 1 |
| | Shaft Layer 3 | 624×38×38 | $1\times1, 624$ conv $3\times3, 624$ conv, stride 1, padding 1 |
| | Detection Layers | - | - |

ratios of the default boxes only needs to consider a few feature values.

## IV. The Proposed Model

In this section we first introduce the architecture of the FSD and give the specific details of the FSN and dense multi-scale features. Next, we show the design of stem and transition layer in backbone. Finally, give the network configurations for the training.

### 1. Model architecture

We present the structure of FSD in Fig. 7, which consists of three parts: backbone, FSN and detection layers. The backbone for feature extraction is mainly composed of a stem, a dense block and a transition layer in series. To make the model train from scratch, we use dense connectivity between each of the shaft layers in the dense block and FSN. Then, all activation functions in FSD are Swish[6] which is defined as $f(x)=x\cdot sigmoid(\beta x)$.

### 2. Feature skyscraper network

The feature skyscraper network (FSN) is composed of a series of shaft layers connected by dense connectivity. The shaft layer has a convolutional layer of $1\times1$ and $3\times3$, and there is a batch normalization (BN) layer and a Swish activation layer before each convolution. As shown in Fig. 7, Each input and output of the shaft layer is concatenated together as input to the next shaft layer. After concatenated, the feature map for each layer is sent to the detect layer.

The resolution $r\times r$ of the feature map depends on the image resolution $R\times R$ of the input model and the minimum side length ($W_b$ or $H_b$) of the bounding box. So, there are: $r=[R/\min(W_b,H_b)]$.

To alleviate the channel accumulation caused by feature reuse, we design a dynamic growth rate k. The growth rate k of the i-th shaft layer is $(L-i+1)\times C$, where L represents the total number of shaft layers, and C represents the number of output channels of the L-th

Table 4 Comparison of FSD with some state-of-the-art models

| Model | Backbone | Pre-train | SPEED(fps) | #Parameters (MB) | mAP | Stem end | Blossom end | Black spot | Input |
|---|---|---|---|---|---|---|---|---|---|
| SSD | Darknet19 | ✓ | 103.038 | 89.419 | 66.837 | 92.364 | 75.051 | 33.097 | 300 |
| SSD | Darknet53 | ✓ | 137.824 | 168.753 | 70.413 | 92.78 | 79.944 | 38.516 | 300 |
| SSD | MobilenetV1 | ✓ | 50.966 | 21.106 | 60.215 | 90.14 | 66.884 | 23.62 | 300 |
| SSD | MobilenetV2 | ✓ | 94.367 | 12.624 | 66.689 | 93.457 | 72.626 | 33.983 | 300 |
| SSD | Resnet50 | ✓ | 99.237 | 46.494 | 68.06 | 92.148 | 75.783 | 36.25 | 300 |
| SSD | VGG16 | ✓ | 95.341 | 91.896 | 69.569 | 89.418 | 77.708 | 41.579 | 300 |
| RFB | Darknet19 | ✓ | 163.490 | 118.798 | 69.757 | 91.287 | 78.239 | 39.745 | 300 |
| RFB | Darknet53 | ✓ | 118.975 | 198.132 | 74.115 | 95.397 | 79.96 | 46.988 | 300 |
| RFB | MobilenetV1 | ✓ | 164.133 | 33.446 | 66.621 | 91.964 | 74.172 | 33.728 | 300 |
| RFB | MobilenetV2 | ✓ | 157.720 | 14.142 | 69.179 | 91.438 | 77.566 | 38.532 | 300 |
| RFB | Resnet50 | ✓ | 105.675 | 75.872 | 72.106 | 94.927 | 77.694 | 43.697 | 300 |
| RFB | VGG16 | ✓ | 77.897 | 121.274 | 74.245 | 94.908 | 80.097 | 47.732 | 300 |
| FSSD | Darknet19 | ✓ | 105.265 | 129.558 | 76.731 | 95.8 | 88.474 | 45.917 | 300 |
| FSSD | Darknet53 | ✓ | 94.267 | 208.892 | 78.41 | 95.088 | 91.883 | 48.261 | 300 |
| FSSD | MobilenetV1 | ✓ | 118.831 | 66.231 | 77.931 | 94.204 | 85.145 | 54.445 | 300 |
| FSSD | MobilenetV2 | ✓ | 134.620 | 22.002 | 77.568 | 96.977 | 83.941 | 51.786 | 300 |
| FSSD | VGG16 | ✓ | 77.073 | 122.524 | 73.16 | 96.087 | 86.201 | 37.193 | 300 |
| YOLO v2 | Darknet19 | ✓ | 150.250 | 193.002 | 80.509 | 94.274 | 86.663 | 60.59 | 416 |
| YOLO v2 | MobilenetV1 | ✓ | 148.160 | 129.674 | 80.175 | 94.911 | 86.574 | 59.04 | 416 |
| YOLO v2 | MobilenetV2 | ✓ | 127.692 | 99.563 | 58.668 | 78.597 | 54.166 | 43.239 | 416 |
| YOLO v3 | Darknet53 | ✓ | 82.577 | 235.628 | 86.458 | 94.867 | 93.71 | 70.798 | 416 |
| YOLO v3 | MobilenetV1 | ✓ | 118.625 | 92.967 | 86.317 | 97.388 | 90.069 | 71.494 | 416 |
| YOLO v3 | MobilenetV2 | ✓ | 111.549 | 85.816 | 85.406 | 93.05 | 92.018 | 71.15 | 416 |
| DSOD | DenseNet | ✗ | 78.842 | 49.422 | 72.298 | 92.336 | 81.053 | 43.505 | 300 |
| FSD | - | ✗ | 131.020 | **5.812** | **87.479** | 98.235 | 86.882 | 77.318 | 150 |

shaft layer.

## 3. Dense multi-scale features

Different from the general feature pyramids, the most remarkable feature of our design FSN is that the resolution of feature maps is constant. However, this does not mean losing the characteristics of multi-scale features. In fact, our multi-scale characteristics are implemented by default boxes of different scales for each layer. This means that larger scale feature maps will be denser.

To adapt to the detection target with a large range of size changes, one stage models such as SSD and DSOD adopt six standard scale (0.2~0.9) default boxes. The scale (0.267) of our largest detection object is comparable to the SSD minimum detection object (0.2), so there is no need to set a series of larger scales of default boxes. But our smallest scale is ten times the largest. At the same time, we observe that many black spots are densely distributed in parts of the navel orange and the shape of the stem (blossom) end is more complicated. So, we design a smaller range for multi-scales.

Table 5 SE layer improves model performance. See Table 6 for more details.

|  | FSD | | | |
|---|---|---|---|---|
| Stem(SE Layer) | ✗ | ✓ | ✗ | ✓ |
| Transition(SE Layer) | ✗ | ✗ | ✓ | ✓ |
| mAP | 82.17 | 83.61 | 87.11 | 87.48 |

The multi-scale lower bound is designed as $1/r$, which is the minimum default box ratio corresponding to the feature map that completely covers the original image. And the upper bound is $Max(W_b, H_b)/R$. Within this range, we use k-means clustering to determine the scale of the default box for each layer.

## 4. Backbone of FSD

The dense block in backbone has the same structure as the feature skyscraper, but the growth rate k is fixed.

We added Channel-wise Attention Mechanism in FSD's transition layer and stem block, as shown in Fig. 8. Many models[33, 34, 39, 40] have introduced this method and it can effectively improve the performance. Our SE layer consists of an average pooling layer, two fully connection layers, and two swish activation function layers. And it is added between the convolution layer and the max pooling layer. In the subsequent experiments, it was also confirmed that the addition of the structure also improved the detection accuracy of our model, as shown in Table 5.

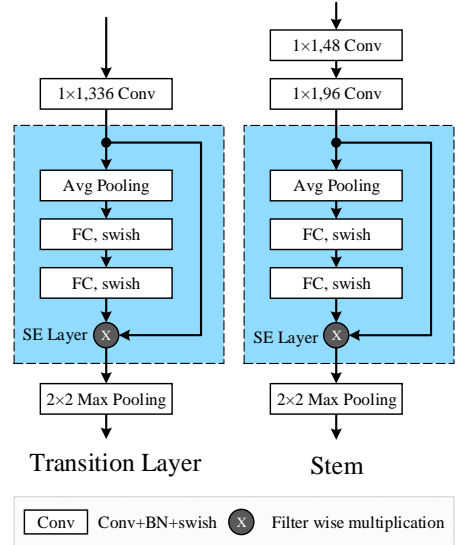

Fig. 8 The transition layer and stem of the FSD architecture. They both add the SE layer structure and use the swish activation function.

## 5. Network configurations

The implementation details of FSD are shown in the Table 3. The resolution of each feature map of the FSN output is $38 \times 38$, and the growth rate k of the 3 shaft layers are: 144, 96 and 48. As shown in Table 2, the average aspect ratios of the bounding boxes in dataset is 1.483, and we set it as the aspect ratios of the default boxes. Then we use K-means to calculate the three cluster centers (They are: 59.75, 165.8, and 349.0.) of the area size as the scale of the default boxes.

Table 6 Verify the FSD module and its design effectiveness. The models in the table are all modified based on FSD, and the input resolution is uniformly set to 150. Where #input channels is the number of channels input to the FSN.

| # Shaft Layers | Non-Linear activations | Stem (SE Layer) | Transition (SE Layer) | SPEED (fps) | #Parameters (MB) | mAP | Stem end | Blossom end | Black spot |
|---|---|---|---|---|---|---|---|---|---|
| 0 | Swish | ✓ | ✓ | 157.958 | 3.176 | 80.658 | 94.303 | 78.427 | 69.243 |
| 1 | Swish | ✓ | ✓ | 143.088 | 3.632 | 84.782 | 94.398 | 86.845 | 73.104 |
| 2 | Swish | ✓ | ✓ | 146.399 | 4.123 | 83.774 | 91.686 | 84.009 | 75.626 |
| 3 | Swish | ✓ | ✓ | 140.976 | 4.650 | 85.799 | 94.132 | 90.507 | 72.757 |
| 4 | Swish | ✓ | ✓ | 127.848 | 5.213 | 86.256 | 95.705 | 90.544 | 72.519 |
| **5** | **Swish** | ✓ | ✓ | **131.020** | **5.812** | **87.479** | **98.235** | **86.882** | **77.318** |
| 6 | Swish | ✓ | ✓ | 126.276 | 6.449 | 85.691 | 96.158 | 85.389 | 75.525 |
| 7 | Swish | ✓ | ✓ | 119.525 | 7.120 | 84.597 | 94.28 | 87.66 | 71.852 |
| 5 | elu | ✓ | ✓ | 129.221 | 5.812 | 84.846 | 95.252 | 85.644 | 73.644 |
| 5 | leaky relu | ✓ | ✓ | 135.198 | 5.812 | 84.804 | 91.176 | 88.362 | 74.874 |
| 5 | prelu | ✓ | ✓ | 131.586 | 5.812 | 85.888 | 96.714 | 86.494 | 74.454 |
| 5 | relu | ✓ | ✓ | 128.031 | 5.812 | 85.318 | 93.067 | 88.787 | 74.101 |
| 5 | selu | ✓ | ✓ | 132.222 | 5.814 | 81.415 | 93.408 | 78.705 | 72.133 |
| 5 | sigmoid | ✓ | ✓ | 126.218 | 5.812 | 81.836 | 90.605 | 82.758 | 72.146 |
| 5 | tanh | ✓ | ✓ | 134.400 | 5.812 | 83.241 | 93.921 | 83.551 | 72.25 |
| 5 | Swish | ✗ | ✓ | 135.267 | 5.787 | 87.111 | 97.528 | 88.257 | 75.548 |
| 5 | Swish | ✓ | ✗ | 129.689 | 5.522 | 83.61 | 93.7 | 87.505 | 69.624 |
| 5 | Swish | ✗ | ✗ | 133.153 | 5.498 | 82.171 | 91.289 | 83.427 | 71.795 |

Table 7 The size of the default boxes is determined based on the clustering center of the bounding boxes in the dataset.

| #Shaft Layers | Cluster Centers | #Parameters (MB) | SPEED (fps) | mAP | Stem end | Blossom end | Black spot |
|---|---|---|---|---|---|---|---|
| 1 | 113.1 | 4.476 | 138.182 | 85.103 | 92.325 | 87.216 | 75.767 |
| 2 | 74.5, 242.1 | 5.325 | 134.971 | 87.252 | 95.89 | 87.768 | 78.401 |
| 3 | 59.7, 165.8, 349.0 | 5.812 | 131.02 | **87.479** | 98.235 | 86.882 | 77.318 |
| 4 | 50.4, 124.7, 220.0, 399.1 | 6.151 | 123.149 | 85.259 | 93.965 | 86.167 | 75.645 |
| 5 | 46.9, 109.5, 185.3, 298.3, 491.5 | 6.502 | 123.993 | 86.854 | 94.841 | 91.898 | 73.824 |
| 6 | 41.0, 86.7, 141.1, 206.5, 309.5, 492.7 | 6.864 | 121.121 | 85.331 | 92.818 | 87.76 | 75.415 |

# V. Experiment

We conduct and evaluate all experiments on our orange dataset benchmark, and present model performance measured by mean Average Precision (mAP), parameters and frames per second (FPS). In Sec 1, we introduce the implementation details of all experiments. In Sec 2, the comparisons with state-of-the-art approaches are performed.

## 1. Implementation details

We implement FSD based on Pytorch framework and trained them from scratch using SGD with initial learning rate 0.01, 0.9 momentum and 0.0001 weight decay on Nvidia TitanX cuDNN v6.0.21 with Intel Xeon E5-2683 v3 @2.00GHz. Then Stochastic Gradient Descent with Restarts (SGDR) makes the learning rate gradually decreases through training, and other training strategies follow SSD. For each scale feature map of the output, we use the same L2 normalization technique as DSOD do. In training, our model inherits the training methods of DSOD and the data augmentation of SSD.

## 2. Comparison with State-of-the-art

We uniformly set the batch size to 8, and the training epoch to 300, which is compared with other state-of-the-art models under our navel orange dataset. In addition to the DSOD and FSD training from scratch, other comparison models are loaded with the best weights of VOC pre-training for training and the default box scale and size are set by default. The training code and detailed parameter settings of the comparison model are available[3].

Intuitively, as shown in Fig. 1, the FSD achieves the highest mAP with the smallest number of parameters, and the inference speed fully meets the real-time requirements. More specifically, as shown in Table 4, FSD has an absolute advantage over most models. It is worth noting that FSD exceeds 1.021% of YOLOv3-Darknet53 with the highest mAP 87.479%, and its parameters is only 2.47% of YOLOv3-Darknet53. And FSD only increased the 15.181% of mAP with 11.76% of the DSOD parameter. Fig. 9 shows the inference results of FSD. It is worth noting that YOLOv3 and FSSD use the FPN-style feature pyramid, and the RFB uses the SSD-style feature pyramid.

# VI. Discussion

On the one hand, we compare the feature skyscraper with several typical feature pyramids and explore the multi-scale feature configuration of the FSN. On the other hand, it is further validated that the effectiveness of the design for the backbone.

---

[3] https://github.com/ShuangXieIrene/ssds.pytorch

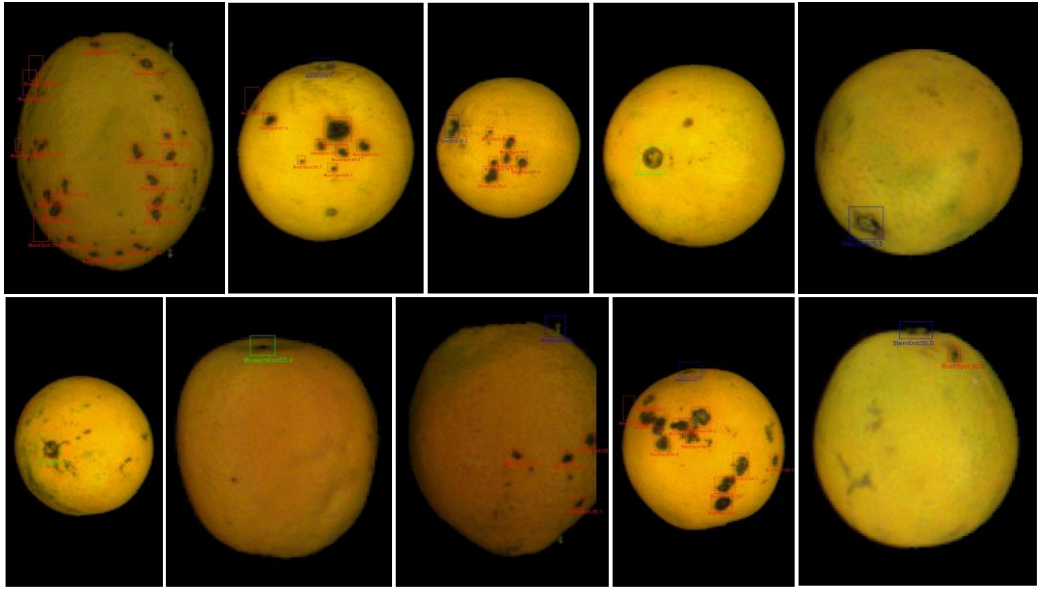

Fig. 9 Show the detection results of FSD on navel orange. The stem end and blossom end can be detected from a bunch of disturbing black spots, and even FSD can identify them at the darker edges.

1. **Feature skyscraper vs. feature pyramids**

To compare the feature skyscraper and feature pyramids, we sequentially replace the FSN with the dsod-style, ssd-style, and FPN-style feature pyramids on the FSD backbone. And three levels (59.7, 165.8, 349.0) are uniformly configured for multi-scale feature maps.

As shown in Table 8, FSN has good results for complex small target stem ends and is on average 4.382% higher than the other three feature pyramids. And black spot is also an average of 3.338% higher. Overall, the mAP of the FSN is an average of 2.973% higher and its speed is slightly superior. The resolution invariance of FSN provides dense feature maps for larger scale default boxes, so the default boxes are effectively covered in the candidate region of the detection object. As shown in the Fig. 3 visualization, most stem ends are indeed distributed near the second level cluster center 165.8.

Table 8 Comparison of feature skyscraper and several feature pyramids models. The backbone is based on FSD.

| Multi-scale Feature | #Parameters (MB) | SPEED (fps) | mAP | Stem end | Blossom end | Black spot |
|---|---|---|---|---|---|---|
| FSN | 5.812 | 131.02 | **87.479** | 98.235 | 86.882 | 77.318 |
| DSOD | 5.637 | 125.761 | 84.957 | 92.482 | 87.513 | 74.877 |
| FPN | 7.950 | 124.371 | 84.451 | 93.873 | 84.881 | 74.598 |
| SSD | 7.450 | 130.782 | 84.109 | 95.204 | 84.657 | 72.465 |

2. **Multi-scale default box**

The resolution of feature maps for DSOD, SSD and FPN is reduced as the scale is reduced, which constrains the minimum effective setting size of the default box. That is because low resolution feature map corresponds to large size default boxes, which cannot effectively set small values for smaller targets. However, FSN ensures that there is more room for the default box scale setting, which is especially effective when the detection object is generally small. As shown in Table 7, we explore the performance of models at different default boxes. The feature maps of FSD has resolution invariance, but different levels of resolution allocate default boxes of different scales. The mAP of multi-scale default boxes is generally higher than that of a single scale. In particular, setting the cluster center to 3 is the best.

3. **Model improvement**

**Feature extraction.** In the first part of the Table 6, retaining a shaft layer as a dense block can achieve a 4.124% improvement (compared to no shaft layer) in model performance. So, we confirm that it is necessary to design at least one shaft layer as a dense block, and the shaft layer is set to 5 layers to make the mAP achieve better results and have a reasonable #Parameters. And increasing the number of shaft layer can effectively improve the feature extraction ability of the back bone, but this is limited.

**Activation functions.** In the second part of the Table 6, we replaced the activation function in the architecture and found that the effect of the Swish activation function is optimal. By replacing the commonly used ReLU activation function with Swish, the overall performance of the model is improved by 2.161%.

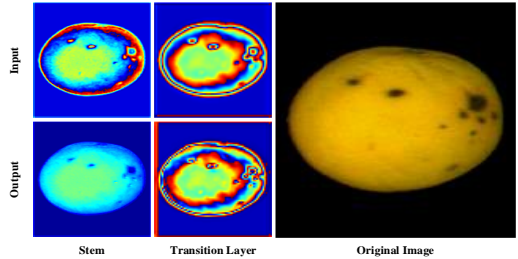

Fig. 10 Visualize the results before and after the SE Layer input. The SE Layer in stem reduces the difference in the input feature map for each channel. In contrast, the SE Layer in the transition layer increases the difference.

**Attention mechanism.** In the third part of the Table 6, we verified the validity of adding the SE Layer. It is proved in the experiment that adding the SE layer to the stem and the transition layer respectively not only improves the model, but both adding them also makes the model further improved. We visualized the feature map before inputting the SE Layer and after outputting the SE layer, in the Fig. 10.

## VII. Conclusion

In this work, to accurately and efficiently distinguish the stem end and the blossom end of navel orange from its black spot, we built a navel orange dataset and proposed the feature skyscraper detector (FSD) model based on the dataset statistical properties. Different from the general feature pyramids, the feature skyscraper network (FSN), the main part of the model, is proposed for multi-scale small objects, and combined with the design of multi-scale default boxes to generate dense multi-scale feature maps. Adding the attention mechanism and application of dense connectivity and Swish in the backbone to further improve the performance of the model.

In the experiment, FSD achieved mAP of 87.479% at 131FPS with only 5.812M parameters, and which mAP exceeds 15.181% of the DSOD that also applies dense connectivity. It is competitive with the state-of-the-art one-stage detectors like SSD, DSOD, YOLOv2, YOLOv3, RFB and FSSD with high detection accuracy, most compact architecture, and real-time performance. And we verify that FSN is better than the general feature pyramids under our data set. In particular, the complexity of small object stem end reached 98.235% (blossom end reached 86.882%) in FSD, and the centralized distribution of small object black spot reached 77.318%.

## References

[1] H. J. Korf. Survival of Phyllosticta citricarpa, anamorph of the citrus black spot pathogen. University of Pretoria, 1998.
[2] A. Kamilaris and F. Prenafeta-Boldu. Deep learning in agriculture: A survey. *Computers and Electronics in Agriculture*, vol. 147, pp. 70-90, 2018.
[3] Z. Shen, Z. Liu, J. Li, Y.-G. Jiang, Y. Chen and X. Xue. Dsod: Learning deeply supervised object detectors from scratch. In *Proceedings of the IEEE International Conference on Computer Vision*, pp. 1919-1927, 2017.
[4] G. Huang, Z. Liu, L. van der Maaten, K. Q. Weinberger and Ieee. Densely Connected Convolutional Networks. In *30th IEEE/CVF Conference on Computer Vision and Pattern Recognition (CVPR)*, Honolulu, HI, pp. 2261-2269, 2017.
[5] J. Hu, L. Shen and G. Sun. Squeeze-and-Excitation Networks. *ArXiv e-prints*, vol. 1709, 2017.
[6] P. Ramachandran, B. Zoph and Q. V. Le. Searching for activation functions. *arXiv preprint arXiv:1710.05941*, 2017.
[7] W. Liu et al. Ssd: Single shot multibox detector. In *European conference on computer vision*, Springer, pp. 21-37, 2016.
[8] J. Redmon and A. Farhadi. YOLO9000: better, faster, stronger. *arXiv preprint*, 2017.
[9] J. Redmon and A. Farhadi. Yolov3: An incremental improvement. *arXiv preprint arXiv:1804.02767*, 2018.
[10] S. Liu and D. Huang. Receptive field block net for accurate and fast object detection. In *Proceedings of the European Conference on Computer Vision (ECCV)*, pp. 385-400, 2018.
[11] Z. Li and F. Zhou. FSSD: feature fusion single shot multibox detector. *arXiv preprint arXiv:1712.00960*, 2017.
[12] A. G. Howard et al. Mobilenets: Efficient convolutional neural networks for mobile vision applications. *arXiv preprint arXiv:1704.04861*, 2017.
[13] M. Sandler, A. Howard, M. Zhu, A. Zhmoginov and L.-C. Chen. Mobilenetv2: Inverted residuals and linear bottlenecks. In *Proceedings of the IEEE Conference on Computer Vision and Pattern Recognition*, pp. 4510-4520, 2018.
[14] K. He, X. Zhang, S. Ren, J. Sun and Ieee. Deep Residual Learning for Image Recognition. In *2016 IEEE Conference on Computer Vision and Pattern Recognition (CVPR)*, Seattle, WA, pp. 770-778, 2016.
[15] K. Simonyan and A. Zisserman. Very deep convolutional networks for large-scale image recognition. *arXiv preprint arXiv:1409.1556*, 2014.
[16] S. K. Behera, L. Jena, A. K. Rath and P. K. Sethy. Disease Classification and Grading of Orange Using Machine Learning and Fuzzy Logic. In *2018 International Conference on Communication and Signal Processing (ICCSP)*, IEEE, pp. 0678-0682, 2018.
[17] D. Rong, Y. Ying and X. Rao. Embedded vision detection of defective orange by fast adaptive lightness correction algorithm. *Computers and Electronics in Agriculture*, vol. 138, pp. 48-59, 2017.
[18] D. Zhang, K. D. Lillywhite, D.-J. Lee and B. J. Tippetts. Automated apple stem end and calyx detection using evolution-constructed features. *Journal of Food Engineering*, vol. 119, no. 3, pp. 411-418, 2013.
[19] L. Liu et al. Deep learning for generic object detection: A survey. *arXiv preprint arXiv:1809.02165*, 2018.
[20] S. Agarwal, J. O. D. Terrail and F. Jurie. Recent Advances in Object Detection in the Age of Deep Convolutional Neural Networks. *arXiv preprint arXiv:1809.03193*, 2018.
[21] R. Girshick, J. Donahue, T. Darrell and J. Malik. Rich feature hierarchies for accurate object detection and semantic segmentation. In *Proceedings of the IEEE*

*conference on computer vision and pattern recognition*, pp. 580-587, 2014.

[22] R. Girshick. Fast r-cnn. In *Proceedings of the IEEE international conference on computer vision*, pp. 1440-1448, 2015.

[23] S. Ren, K. He, R. Girshick and J. Sun. Faster r-cnn: Towards real-time object detection with region proposal networks. In *Advances in neural information processing systems*, pp. 91-99, 2015.

[24] J. Dai, Y. Li, K. He and J. Sun. R-fcn: Object detection via region-based fully convolutional networks. In *Advances in neural information processing systems*, pp. 379-387, 2016.

[25] J. Redmon, S. Divvala, R. Girshick and A. Farhadi. You only look once: Unified, real-time object detection. In *Proceedings of the IEEE conference on computer vision and pattern recognition*, pp. 779-788, 2016.

[26] J. Huang et al. Speed/accuracy trade-offs for modern convolutional object detectors. In *Proceedings of the IEEE conference on computer vision and pattern recognition*, pp. 7310-7311, 2017.

[27] A. Krizhevsky, I. Sutskever and G. E. Hinton. ImageNet Classification with Deep Convolutional Neural Networks. *Communications of the Acm*, vol. 60, no. 6, pp. 84-90, 2017.

[28] S. Ioffe and C. Szegedy. Batch Normalization: Accelerating Deep Network Training by Reducing Internal Covariate Shift. *ArXiv e-prints*, vol. 1502, 2015.

[29] C. Szegedy, V. Vanhoucke, S. Ioffe, J. Shlens and Z. Wojna. Rethinking the Inception Architecture for Computer Vision. *ArXiv e-prints*, vol. 1512, 2015.

[30] C. Szegedy, S. Ioffe, V. Vanhoucke and A. Alemi. Inception-v4, Inception-ResNet and the Impact of Residual Connections on Learning. *ArXiv e-prints*, vol. 1602, 2016.

[31] O. Russakovsky et al. Imagenet large scale visual recognition challenge. *International journal of computer vision*, vol. 115, no. 3, pp. 211-252, 2015.

[32] T.-Y. Lin, P. Dollár, R. Girshick, K. He, B. Hariharan and S. Belongie. Feature pyramid networks for object detection. In *Proceedings of the IEEE Conference on Computer Vision and Pattern Recognition*, pp. 2117-2125, 2017.

[33] Q. Zhao et al. M2Det: A Single-Shot Object Detector based on Multi-Level Feature Pyramid Network. *arXiv preprint arXiv:1811.04533*, 2018.

[34] Y. Yang, Z. Zhong, T. Shen and Z. Lin. Convolutional Neural Networks with Alternately Updated Clique. *ArXiv e-prints*, vol. 1802, 2018.

[35] X. Glorot, A. Bordes and Y. Bengio. Deep sparse rectifier neural networks. In *Proceedings of the fourteenth international conference on artificial intelligence and statistics*, pp. 315-323, 2011.

[36] K. He, X. Zhang, S. Ren and J. Sun. Delving deep into rectifiers: Surpassing human-level performance on imagenet classification. In *Proceedings of the IEEE international conference on computer vision*, pp. 1026-1034, 2015.

[37] D.-A. Clevert, T. Unterthiner and S. Hochreiter. Fast and accurate deep network learning by exponential linear units (elus). *arXiv preprint arXiv:1511.07289*, 2015.

[38] G. Klambauer, T. Unterthiner, A. Mayr and S. Hochreiter. Self-normalizing neural networks. In *Advances in neural information processing systems*, pp. 971-980, 2017.

[39] L. Chen et al. Sca-cnn: Spatial and channel-wise attention in convolutional networks for image captioning. In *Proceedings of the IEEE conference on computer vision and pattern recognition*, pp. 5659-5667, 2017.

[40] F. Wang et al. Residual Attention Network for Image Classification. *30th Ieee Conference on Computer Vision and Pattern Recognition*, pp. 6450-6458, 2017.